\documentclass{article}
\usepackage{arxiv}
\usepackage{amsmath,amsfonts}

\newcommand{\revised}[1]{#1}

\usepackage{algorithm}
\usepackage{algorithmic}
\usepackage{array}
\usepackage[font=footnotesize,labelfont=sf,textfont=sf]{subfig}
\usepackage{textcomp}
\usepackage{url}
\usepackage{verbatim}
\usepackage{graphicx}
\usepackage[nolist]{acronym}
\usepackage{booktabs}
\usepackage{etoolbox}
\usepackage{float}
\usepackage{tikz}
\usetikzlibrary{arrows.meta,positioning}

\renewcommand{\theparagraph}{\ifnum\value{subsubsection}=0\thesubsection\else\thesubsubsection\fi\alph{paragraph}}

\def\BibTeX{{\rm B\kern-.05em{\sc i\kern-.025em b}\kern-.08em
    T\kern-.1667em\lower.7ex\hbox{E}\kern-.125emX}}

\begin{document}

\begin{acronym}
  \acro{AI}{Artificial Intelligence}
  \acro{ML}{Machine Learning}
  \acro{DL}{Deep Learning}
  \acro{IoT}{Internet of Things}
  \acro{IIoT}{Industrial Internet of Things}
  \acro{CM}{Condition Monitoring}
  \acro{HI}{Health Index}
  \acro{RUL}{Remaining Useful Life}
  \acro{PdM}{Predictive Maintenance}
  \acro{PvM}{Preventive Maintenance}
  \acro{RM}{Reactive Maintenance}
  \acro{SSM}{State Space Models}
  \acro{SQR}{Simultaneous Quantile Regression}
  \acro{PHM}{Prognostics and Health Management}
  \acro{PHM18}{2018 PHM Data Challenge}
  \acro{UB}{Unexpected Breaks}
  \acro{UL}{Unexploited Lifetime}
  \acro{PINN}{Physics-Informed Neural Network}
  \acro{BNN}{Bayesian Neural Network}
  \acro{MLP}{Multilayer Perceptron}
  \acro{CNN}{Convolutional Neural Network}
  \acro{LSTM}{Long Short-Term Memory}
  \acro{RNN}{Recurrent Neural Network}
  \acro{GRU}{Gated Recurrent Unit}
  \acro{S4}{Structured State Spaces}
  \acro{S4D}{Diagonal State Spaces}
  \acro{S5}{Simplified Structured State Spaces}
  \acro{TCN}{Temporal Convolutional Network}
  \acro{QR}{Quantile Regression}
  \acro{MSE}{Mean Squared Error}
  \acro{RMSE}{Root Mean Squared Error}
  \acro{MAE}{Mean Absolute Error}
  \acro{CDF}{Cumulative Distribution Function}
  \acro{IBE}{Ion Beam Etching}
  \acro{IME}{Ion Milling Equipment}
  \acro{PBN}{Particle Beam Neutralizer}
  \acro{FCP}{Flowcool Pressure Dropped Below Limit}
\end{acronym}


\title{
  Uncertainty and Business-Aware Remaining Useful Life Estimation for Semiconductor Manufacturing
  \thanks{Corresponding author: davide.frizzo.1@studenti.unipd.it}
  \thanks{This work was funded by the European Union in the context of the Horizon Europe project ``AIMS5.0---Artificial Intelligence in Manufacturing Leading to Sustainability and Industry 5.0''. Grant Agreement ID:\ 101112089.}
}

\author{
  Davide Frizzo, Francesco Borsatti, and Gian Antonio Susto \\
  Department of Information Engineering, University of Padova
}

\maketitle

\begin{abstract}


\revised{\ac{RUL} estimates guide maintenance in semiconductor manufacturing, where a
late intervention can disrupt production but an early one discards usable
equipment life. A single prediction does not express the uncertainty needed to
choose between these outcomes. We combine \ac{DL} sequence models with
\ac{SQR} to estimate \ac{RUL} at selectable quantile levels, and evaluate
maintenance decisions using \ac{UB}, \ac{UL}, and a cost-weighted objective.
On the $F_1$ fault mode of the \ac{PHM18} ion-milling dataset, we compare
recurrent, Transformer, and \ac{SSM} architectures. \ac{S4D} gives the
strongest overall \ac{RUL} accuracy across the evaluated quantiles. Varying
the quantile and maintenance window exposes the trade-off between failure risk
and early intervention; under the examined cost ratios, the resulting decisions
have lower cost than the \ac{PvM} baselines.}

\end{abstract}



\section{Introduction} \label{sec:introduction}



\revised{Semiconductor manufacturing depends on interconnected equipment, so the failure
of a single asset can interrupt production. Identifying the components most
likely to fail in the near term is therefore essential to enable timely
maintenance, preserve pipeline continuity, and avoid costly
repairs~\cite{semiconductor_cost}. Traditional approaches include \ac{RM} and
\ac{PvM}, in which maintenance is performed after failure or at fixed
intervals, respectively. On the other hand, \ac{IoT}
sensors~\cite{iot_pdm,iot_pdm_2} provide data that \ac{AI}, \ac{ML}, and
\ac{DL} models can use for industrial
prediction~\cite{iot_survey,industry_4_ml_survey}. In \ac{PdM}, maintenance
decisions based on these predictions must balance two consequences: intervening
too late can cause \ac{UB}, while intervening too early leaves useful equipment
life unexploited (\ac{UL}).}

\revised{A point estimate of \ac{RUL} does not show how uncertain that estimate is.
This matters because overestimating and underestimating \ac{RUL} have different
operational costs: prediction accuracy alone cannot determine when to
intervene. Predictive quantiles allow maintenance decisions to be examined at
different levels of conservatism, while business metrics measure the resulting
breaks and unexploited life. We therefore combine \ac{DL} sequence models with
\ac{SQR}~\cite{sqr} and evaluate their predictions and maintenance decisions
on ion-milling data from the \ac{PHM18} dataset.}



\revised{The main contributions of this work are as follows:

\begin{itemize}
    \item We integrate sequence-learning models for \ac{RUL} estimation with
    \ac{SQR} to obtain predictive quantiles that characterize uncertainty.
    \item We assess maintenance decisions using business-oriented metrics that
    quantify unexpected breaks and unexploited life, enabling the selection of
    a maintenance strategy for a given cost trade-off.
    \item We benchmark different model architectures on the $F_1$ fault mode of
    the \ac{PHM18} dataset, comparing prediction accuracy, computational cost
    and business metrics.
    \item We make the code used to produce the reported results publicly
    available at \url{https://github.com/AMCO-UniPD/ssm-pdm}.
\end{itemize}}

The paper is organized as follows. Section~\ref{sec:related-works} provides
background on \ac{RUL} estimation, and Section~\ref{sec:proposed-method}
describes the proposed approach. Section~\ref{sec:exp-setup} presents the
experimental setup, including the dataset description and preprocessing steps,
and Section~\ref{sec:exp-results} reports the experimental evaluation. Finally,
Section~\ref{sec:conclusion} discusses conclusions and future work.


\section{Related Work}\label{sec:related-works}


The aim of \ac{PdM} is to model machine degradation by using signals from
multiple \ac{CM} sensors as inputs and producing a so-called \ac{HI} that
indicates when the equipment reaches a low health level and requires
maintenance. When this threshold is reached, the machine raises an alarm and
maintenance scheduling begins. One of the most commonly used health indicators
is \ac{RUL}, a decreasing signal that represents the amount of useful life left
in the equipment. Consequently, \ac{PdM} is normally framed as a \ac{RUL}
estimation problem.




Different kinds of predictive models can be used to estimate \ac{RUL}.
Traditionally, model-based approaches that represent the physics of the
equipment under analysis were employed~\cite{model_based_bearing}. Their
knowledge of the system dynamics enables accurate simulation even when the
system configuration changes. However, their computational requirements and
deployment costs often prevent economical scaling. Owing to advances
in \ac{AI}, most recent \ac{RUL} frameworks instead use data-driven approaches.
These methods exploit the large volumes of data produced by \ac{CM} sensors to
model relationships between sensor measurements and machine degradation. They
can produce \ac{RUL} estimates within fractions of a second and are therefore
more readily deployed on devices with limited memory and compute capabilities.
Their main limitation is the lack of physical knowledge, which can reduce their
effectiveness when the machine configuration changes. Consequently, hybrid
models, usually based on \ac{PINN}s, have recently been
developed~\cite{pinn,hybrid_pdm_olga}. This study focuses on data-driven \ac{RUL}
estimation, for which both \ac{ML} and \ac{DL} based approaches can be used.
\ac{ML} models generally require an additional preprocessing step to extract
meaningful features from the raw
signals~\cite{feat_ext_tutorial}. These typically include statistical (e.g.,
mean, root mean square, skewness, kurtosis, \dots), temporal, and frequency-domain
(e.g., power spectrum) features, which are computed over sliding windows to
better highlight the equipment's life degradation trend. The enlarged feature
set is then normally used to train \ac{ML} models~\cite{rul_sensor_fusion} but
can also provide additional context to \ac{DL}-based
approaches~\cite{ind_ceruleo}. \ac{DL} models are much more expressive than
\ac{ML} architectures, and their hierarchical structure, made up of multiple
hidden layers, enables automatic extraction of the features needed for accurate
\ac{RUL} estimation. This comes at the cost of larger models
in terms of memory and compute. Moreover, these models normally require a
significant amount of data to achieve optimal performance. Nevertheless,
modern \ac{IIoT} systems produce large volumes of data, enabling the
development of numerous \ac{DL}-based approaches to \ac{RUL}
estimation~\cite{RUL_CMAPSS}. Common architectures include \ac{LSTM}s
~\cite{cmapss_bidir_lstm}, \ac{RNN}s~\cite{phm_RNN_online}, \ac{CNN}-\ac{LSTM}
combinations~\cite{cmapss_stf}, and attention layers~\cite{transformer_dual},
which model multivariate \ac{IoT} time series.


The \ac{RUL} estimation task is not new in the context of semiconductor
manufacturing. It gained popularity through \ac{PHM18}~\cite{PHM2018}, which
introduced a high-quality, challenging dataset on ion-milling machines, an
important type of equipment in chip fabrication that requires careful
monitoring. Several studies, including this one, have used this data source to
train and validate their \ac{PdM} approaches for wafer
fabrication~\cite{pdm_semiconductor_survey}. The authors of~\cite{phm_tcn_lstm}
combine a \ac{TCN}, an \ac{LSTM}, and attention to predict \ac{RUL} on
\ac{PHM18}. The study in~\cite{phm_transfer_learning} addresses an important
problem in \ac{PHM}: varying machine operating conditions and fault types. In
particular, a \ac{TCN}-based model is trained on the base fault in \ac{PHM18}
and then fine-tuned on other fault types. Finally,~\cite{phm_transformer}
considers a multiscale Transformer~\cite{transformer}. An \ac{LSTM} layer is
integrated into the attention block to combine its features with those
extracted by the Transformer through a Hadamard product. Although \ac{DL}-based
approaches can accurately determine the end of a machine's life, their
predictions involve uncertainty arising both from the data (i.e., aleatoric
uncertainty) and from the model itself (i.e., epistemic uncertainty).
Quantifying uncertainty in \ac{PHM} is crucial because a missed maintenance
intervention can have catastrophic economic consequences. Several approaches to
uncertainty estimation exist in the \ac{DL}
literature~\cite{uncertainty_evaluation}; most are based on Bayesian inference
and \ac{BNN}s~\cite{uncertainty_bayesian}. More recently, Quantile Regression
approaches have been introduced as lightweight, efficient alternatives to
\ac{BNN}s~\cite{pdm_quantile_reg, quantile_rul_perspective}. \revised{In this study,
uncertainty is estimated through \ac{SQR}~\cite{sqr}, which focuses on
aleatoric uncertainty by estimating the quantiles of the posterior predictive
distribution, as detailed in Section~\ref{subsec:quantile-reg}.}


\section{Proposed Method} \label{sec:proposed-method}

\revised{This section presents the proposed \ac{PdM} framework for uncertainty-aware
\ac{RUL} estimation. First, sensor histories are divided into run-to-failure
cycles and assigned piecewise-linear \ac{RUL} targets
(Section~\ref{subsec:rul-estimation}). Each cycle is split into windows, and a
sequence model maps the concatenation of sensor readings and a sampled quantile
level to \ac{RUL} predictions. A weighted loss gives greater emphasis to the
windows near failure (Section~\ref{subsec:loss-function}), while \ac{SQR}
produces estimates at different quantile levels to express predictive
uncertainty (Section~\ref{subsec:quantile-reg}). At inference time, the window
predictions are joined into a \ac{RUL} trajectory. A maintenance notification
is triggered when that trajectory crosses a selected safety threshold. We
evaluate the resulting decisions, at different quantile levels, using business
metrics such as \ac{UB}, \ac{UL}, and their cost-weighted combination $J$
(Section~\ref{subsec:business-metrics}). Figure~\ref{fig:method-pipeline}
summarizes the complete pipeline.}


\begin{figure}[H]
  \centering
  \begin{tikzpicture}[
    >=Latex,
    node distance=2.5mm and 5mm,
    pipelinebox/.style={draw, rounded corners=2pt, align=center,
      text width=0.42\columnwidth, minimum height=10mm,
      inner sep=2pt, font=\footnotesize},
    flow/.style={-Latex, thick}
  ]
    \node[pipelinebox] (cycles) {Sensor histories split into\\run-to-failure cycles};
    \node[pipelinebox, right=of cycles] (windows) {Windowing and\\piecewise-linear \ac{RUL}\\targets (training)};
    \node[pipelinebox, below=of windows] (model) {Quantile-conditioned\\sequence model trained\\with weighted pinball loss};
    \node[pipelinebox, left=of model] (trajectory) {Predicted quantile\\\ac{RUL} trajectory};
    \node[pipelinebox, below=of trajectory] (decision) {Maintenance notification\\at predicted \ac{RUL}\\threshold $m$};
    \node[pipelinebox, right=of decision] (evaluation) {Decision evaluation:\\\ac{UB}, \ac{UL}, and cost $J$};
    \draw[flow] (cycles.east) -- (windows.west);
    \draw[flow] (windows.south) -- (model.north);
    \draw[flow] (model.west) -- (trajectory.east);
    \draw[flow] (trajectory.south) -- (decision.north);
    \draw[flow] (decision.east) -- (evaluation.west);
  \end{tikzpicture}
  \caption{Overview of the proposed uncertainty-aware \ac{PdM} pipeline.
    Targets and the weighted loss are used during training; at inference,
    sensor windows and a chosen quantile level produce the \ac{RUL} trajectory.}
  \label{fig:method-pipeline}
\end{figure}

\subsection{\ac{RUL} Estimation} \label{subsec:rul-estimation}

A \ac{RUL} model receives sensor signals from monitored equipment. Data are
split into intervals between maintenance interventions, denoted as
\textit{run-to-failure cycles} \revised{or \textit{lives}}; each cycle
$i\in\{1,\dots,N\}$ is processed independently. Sensor readings are
$\mathcal{X}\in\mathbb{R}^{n\times m}$, where $n$ is the number of samples and
$m$ is the number of sensors. The unmeasurable \ac{RUL} target is
$\mathcal{Y}\in\mathbb{R}^{n}$ and is defined by a degradation model, most
commonly linear or piecewise linear. The linear degradation model is the
simplest of the two and defines the target signal as follows:

\begin{equation} \label{eq:linear-rul}
  \text{RUL} = \{n-1, n-2, \dots, 0\}.
\end{equation}

This model assumes that the machine's health begins to degrade as soon as it
starts operating and continues to degrade until the final sample, at which
point the health status reaches zero and the equipment reaches the end of its
life. By contrast, the piecewise-linear model is defined as $\text{RUL} = \{
\text{MAX\_{RUL}}, \dots, \text{MAX\_{RUL}}, \text{MAX\_{RUL}} - 1, \dots, 0
\}$. The health status starts at $\text{MAX\_{RUL}}$, remains constant for
$n-\text{MAX\_{RUL}}$ samples, and then decreases linearly towards zero as
specified by~\eqref{eq:linear-rul}. This model is generally preferred because it is
reasonable to assume that the machine remains fully healthy for some time
before declining towards failure. The value
$\text{MAX\_{RUL}}$, also known as the \textit{elbow} or \textit{knee} point,
is a hyperparameter that can be defined using reasonable assumptions about
machine life-cycle durations or statistics computed from the training
data~\cite{echo_rul}. We use a piecewise-linear degradation model with
$\text{MAX\_{RUL}} = 500$, following~\cite{phm_RNN_online,phm_transformer}.
Once the target signal is defined, \ac{RUL} estimation can be framed as a
regression task on multivariate time-series data. The learning objective is
therefore to find a function $f: \mathcal{X} \rightarrow \mathcal{Y}$ such
that $y_i \approx f(x_i)$ for every paired sample $(x_i,y_i)$.

%

\revised{We follow the architecture design of~\cite{pdm_quantile_reg} consisting of a
decoder, a backbone and a decoder. The encoder projects each window of sensor
readings into a latent feature sequence. A sequence-learning backbone models
dependencies across time and sensors; we compare \ac{RNN}, \ac{GRU}, \ac{LSTM},
Transformer, \ac{S4}, \ac{S4D}, and \ac{S5} backbones. A linear decoder maps
the resulting features to one \ac{RUL} prediction per time step.
Section~\ref{subsec:quantile-reg} and Algorithm~\ref{alg:quantile-regression}
describe how the quantile level conditions these predictions.}

\subsection{Weighted Loss Function} \label{subsec:loss-function}

Given the long life-cycle durations in the \ac{PHM18} dataset
(Section~\ref{subsec:phm-dataset}), it is impractical to feed an entire signal
to the model. Therefore, similarly to~\cite{pdm_quantile_reg}, we use a
sliding-window approach to divide inputs into non-overlapping subsequences on
which predictions are performed. In particular, this approach divides sequences
into mini-batches that are processed in parallel during training and
validation. Zero-padding ensures that all windows share the same length. The
model outputs the \ac{RUL} signal for an input window and the predictions for
the windows representing a run-to-failure cycle are concatenated to obtain the
final estimate for the entire cycle. Because a piecewise-linear degradation
model is used for \ac{RUL}, the target signal is constant at
$\text{MAX\_{RUL}}$ for most windows. A decreasing trend is observable only in
the windows after the elbow point (i.e., in the final $\text{MAX\_{RUL}}$
samples). During training, the model processes one window at a time; because
most windows have a constant target signal, the model may learn to always
predict a constant \ac{RUL}, making its predictions practically useless. To
address this challenge, which resembles class imbalance in classification
tasks, we employ a weighted loss function. Consider a generic loss function
$\mathcal{L}(f(x),\hat{y})$ \revised{(in this study, the
pinball loss, as explained in Section~\ref{subsec:quantile-reg})}, where $f(x)$ is the
model prediction on an input window and $\hat{y}$ is the corresponding \ac{RUL}
target signal. The loss function employed in the proposed approach can be
formalized as follows:

\begin{equation} \label{eq:weighted-loss}
  \begin{cases}
    \frac{1}{W_{constant}} \cdot \mathcal{L}(f(x),\hat{y}) & \text{if} \ \hat{y} \ \text{is constant} \\
    \frac{1}{W_{decreasing}} \cdot \mathcal{L}(f(x),\hat{y}) & \text{otherwise}
  \end{cases},
\end{equation}

where $W_{constant}$ and $W_{decreasing}$ are the numbers of constant and
decreasing windows, respectively. With this definition, errors on decreasing
windows are penalized more heavily than errors on constant windows. Accurate
prediction of the constant region of the \ac{RUL} signal is not the priority,
because the primary aim of a \ac{PdM} framework is to detect the so-called
\textit{maintenance point}, i.e., the time at which a maintenance intervention
is most appropriate.

\subsection{Uncertainty Estimation through Quantile Regression} \label{subsec:quantile-reg}

As stated in Section~\ref{sec:introduction}, quantifying predictive uncertainty
is fundamental to finding the optimal trade-off between overestimating and
underestimating \ac{RUL}, thereby avoiding \ac{UB} while limiting \ac{UL}. In
semiconductor manufacturing, where equipment is operated by specialized
personnel, uncertainty-aware estimates are particularly important for enabling
operators to make informed, risk-aware maintenance decisions. Bayesian
uncertainty methods require priors and computationally costly sampling
procedures, which can be difficult to apply in high-dimensional settings. We
instead use \ac{SQR}~\cite{sqr}, which adds uncertainty quantification to a
regression model with minimal architectural and training changes. \ac{QR}
estimates quantiles of $p(y|x)$, where $x$ denotes sensor readings and $y$
denotes \ac{RUL}. By contrast, classical regression models trained with
\ac{MSE} or \ac{MAE} estimate the mean and median of $p(y|x)$, respectively.
The selected quantile encodes risk preference. High quantiles produce
optimistic \ac{RUL} estimates, favouring production at greater failure risk;
low quantiles are conservative and protect equipment but can waste useful life.
Given the \ac{CDF} of the target variable, $F(y) = P(Y \leq y)$, the
\textbf{quantile function} for a quantile level $\tau \in [0,1]$ is
$F^{-1}(\tau) = \inf \{y \mid F(y) \geq \tau\}$. \revised{To estimate quantile $\tau$ of
$Y$ given $x$, where $x \in \mathbb{R}^n$, the model $y =
\hat{f}_{\tau}(x)$ approximates the \textbf{conditional quantile function}. The
model is trained using the \textit{pinball loss}, which, according
to~\cite{sqr}, is the optimal one for this task:}

\begin{equation} \label{eq:pinball_loss}
\mathcal{L}_{\tau}(y,\hat{y}) =
\begin{cases}
        \tau(y - \hat{y}) \quad \text{if} \ y \geq \hat{y} \\
        (1-\tau)(\hat{y} - y) \quad \text{otherwise}
\end{cases}.
\end{equation}

This training criterion alone limits estimation to the single quantile $\tau$.
To estimate all quantiles of $p(y|x)$, \ac{SQR} augments each input sample $x_i$
with a randomly sampled quantile $\tau_i$, yielding $(x_i,\tau_i)$. The pinball
loss for $\tau_i$ is then used to update the model parameters:
$\mathcal{L}_{\tau_i}(y_i, \hat{y}_i)$. The full training objective is therefore
$\hat{f} \in
\text{argmin}_{f} \frac{1}{n} \sum_{i=1}^{n} \mathbb{E}_ {\tau \sim U(0,1)}
[\mathcal{L}_{\tau}(f(x_i,\tau),y_i)]$, where $U(0,1)$ denotes the uniform
distribution over the interval $[0,1]$. At inference time, a quantile level
$\tau$ is selected and provided to the model together with the sensor readings
$x$ to obtain the corresponding conditional \ac{RUL} estimate
$\hat{f}(x,\tau)$. \revised{Algorithm~\ref{alg:quantile-regression} summarizes the
training and inference steps.}

\begin{algorithm}[t]
\caption{Quantile-conditioned training and inference for \ac{RUL} estimation.}
\label{alg:quantile-regression}
\renewcommand{\algorithmicrequire}{\textbf{Input:}}
\begin{algorithmic}[1]
  \REQUIRE Training pairs $(x_i,y_i)$, model $f_\theta$, learning rate $\eta$
  \FOR{each training pair $(x_i,y_i)$}
    \STATE Sample $\tau_i \sim U(0.1,0.9)$\footnote{In the experiments, we use $U(0.1,0.9)$ instead of $U(0,1)$ to avoid producing overly extreme estimates.}.
    \STATE Form the augmented input $[x_i;\tau_i]$.
    \STATE Predict $\hat y_i = f_\theta([x_i;\tau_i])$.
    \STATE Update $\theta \leftarrow \theta - \eta\nabla_\theta
      \mathcal{L}_{\tau_i}(y_i,\hat y_i)$ using the pinball loss in
      \eqref{eq:pinball_loss}.
  \ENDFOR
  \STATE \textbf{Inference:} Choose $\tau^* \in [0,1]$.
  \STATE Predict $\hat y^* = f_\theta([x^*;\tau^*])$.
\end{algorithmic}
\end{algorithm}

\subsection{\revised{Business Metrics}} \label{subsec:business-metrics}

As described in Section~\ref{subsec:rul-estimation}, the primary objective of
the proposed model is to accurately estimate the target \ac{RUL} signal.
However, predictive accuracy alone is insufficient in practical industrial
scenarios, where the estimated \ac{RUL} must ultimately support maintenance
decision-making. In particular, each prediction is used to determine the
\emph{maintenance point}, i.e., the time instant at which the system generates
a maintenance notification indicating that the asset is approaching the end of
its useful life; this point is denoted by $T_m$. Typically, $T_m$ is defined as
the time at which $\text{RUL}=0$. However, because the
maintenance notification must be issued sufficiently in advance to allow
maintenance planners to schedule the intervention, procure the required
resources, and complete the maintenance operation before a failure
occurs~\cite{maintenance_schedule}, a more practical definition is needed.
Business metrics introduced in~\cite{pdm_classifier,ind_ceruleo} help define
the maintenance point by accounting for the business and economic costs
associated with different types of errors made by the \ac{RUL} estimation
model. From an operational
perspective, prediction errors have different consequences depending on their
direction. An overestimation of the \ac{RUL} postpones the maintenance
notification and may result in the asset failing before maintenance can be
performed, leading to a so-called \ac{UB} (i.e., $\rho_{UB}$). Conversely, an
underestimation anticipates the maintenance notification, reducing the risk of
failure but potentially causing components to be replaced while useful
operating life remains. This situation is termed \ac{UL} (i.e.,
$\rho_{UL}$). To account for this trade-off, a \textit{maintenance window} $m$
is introduced as a safety margin. \revised{Rather than scheduling maintenance at the
predicted end of life, the maintenance notification is brought forward by $m$.
For each test life $i$, let $T_i$ denote its recorded end-of-life time and
$T_i^*$ the end-of-life time estimated from the predicted \ac{RUL} trajectory.
The maintenance point is $T_{m,i}=T_i^*-m$. With $N$ test lives, the
unexpected-break fraction and mean unexploited lifetime are computed as
\begin{equation} \label{eq:business-metrics}
  \begin{aligned}
    \rho_{UB}(m) &= \frac{1}{N}\sum_{i=1}^{N}
      \mathbb{I}\!\left[T_{m,i}\geq T_i\right], \\
    \rho_{UL}(m) &= \frac{1}{N}\sum_{i=1}^{N}
      \max\!\left(T_i-T_{m,i},\,0\right).
  \end{aligned}
\end{equation}
Thus, underestimating \ac{RUL} leads to early maintenance: the remaining
usable lifetime contributes to \ac{UL}, while no \ac{UB} is incurred.
Conversely, if the maintenance point falls at or after the recorded end of
life, it counts as an \ac{UB} and contributes zero to \ac{UL}.
The maintenance window is subtracted from the predicted end of life because a
more preventive approach is generally preferred: an \ac{UB} typically
costs more than not fully exploiting the machine's lifetime.} Increasing $m$
decreases the likelihood of
unexpected failures by bringing forward the maintenance intervention. However, an
excessively conservative value results in maintenance being performed too
early, thereby reducing the utilisation of the asset and increasing operational
costs. The objective of a \ac{PdM} system is therefore not only to minimise the
prediction error, but also to identify the best compromise between \ac{UB} and
\ac{UL}. This trade-off can be evaluated through the business metric $J$
proposed in~\cite{ind_ceruleo}, which combines \ac{UB} and \ac{UL} into a
single score and enables the selection of the optimal maintenance window $m$.
Metric $J$ is defined as follows:

\begin{equation} \label{eq:metric-J}
  J = \rho_{UB} \cdot c_{UB} + \rho_{UL} \cdot c_{UL},
\end{equation}

\revised{where $c_{UB}$ and $c_{UL}$ are hyperparameters representing the costs
associated with \ac{UB} and \ac{UL}, respectively. The values and relative
magnitudes of these costs
depend on the specific application, but in most cases $c_{UB} \gg c_{UL}$.}


\section{Experimental Setup} \label{sec:exp-setup}

This section describes the setup of the evaluation experiments reported in
Section~\ref{sec:exp-results}. Section~\ref{subsec:phm-dataset} presents the
problem domain (Section~\ref{par:ion-mill-etching}) and the \ac{PHM18}
benchmark dataset (Section~\ref{par:data-description}).
\revised{Section~\ref{subsec:data-preprocessing} then describes data preprocessing, and
Section~\ref{subsec:exp-details} provides some numerical details on the
experiments performed.}

\subsection{\ac{PHM} Dataset} \label{subsec:phm-dataset}

The benchmark dataset adopted for this study is \ac{PHM18}, introduced in the
\acl{PHM18} to enable comparisons among \ac{PHM} solutions for semiconductor
manufacturing. It focuses on ion-milling etching tools and faults associated
with their use.

\paragraph{Ion-Milling Etching Process} \label{par:ion-mill-etching}

Ion milling (\ac{IBE}) precisely removes wafer material. A wafer is processed
in a vacuum chamber through a multi-step recipe specifying beam voltage and
current, incidence angle, rotation speed, and duration. An ion source
accelerates inert-gas ions, typically argon, into a collimated beam; impact
ejects surface atoms through sputtering. A rotating, tiltable stage promotes
uniform removal, a shutter blocks the beam until conditions are reached, and
the \ac{PBN} controls the beam profile and charge distribution. Helium-assisted
water cooling prevents damaging temperature increases. Grid and chamber wear
and flowcool leaks can reduce quality, scrap wafers, and cause downtime.
Accurate health-state and \ac{RUL} estimates therefore support maintenance that
avoids failures while preserving availability.

\paragraph{Dataset Overview} \label{par:data-description}

\ac{PHM18} contains readings from 20 etching tools under two operating modes.
Following the challenge split, 15 tools train the models and five test them:
\texttt{01M02, 02M02, 03M01, 04M01, 06M01}. The run-to-failure cycles in the
dataset are computed based on three different failure modes. In particular, the
end of life is defined as the time at which an operator stops the machine to
perform maintenance (i.e., this generally differs from the time at which the
fault is observed):

\begin{itemize}
  \item $F_1$: \ac{FCP} low
  \item $F_2$: \ac{FCP} high
  \item $F_3$: Flowcool leak
\end{itemize}

We consider only $F_1$, the most represented fault type,
following~\cite{phm_transfer_learning}; all subsequent descriptions and results
use this setting. This yields 703 training cycles and five test cycles.
Table~\ref{tab:phm-stats} reports training-cycle statistics used by the
\ac{PvM} baselines.

\begin{table}[t]
\centering
\begin{minipage}[t]{0.54\columnwidth}
\centering
\caption{\normalfont Statistics of the \ac{PHM} training-cycle durations}
\label{tab:phm-stats}
\begin{tabular}{@{}lr@{}}
\toprule
\textbf{Statistic} & \textbf{Value} \\
\midrule
mean           & 60339.20 \\
median         &  9675.50 \\
quantile\_0.1  &  1054.80 \\
quantile\_0.25 &  3366.75 \\
quantile\_0.75 & 27730.20 \\
quantile\_0.9  & 99975.30 \\
\bottomrule
\end{tabular}
\end{minipage}%
\hfill
\begin{minipage}[t]{0.43\columnwidth}
\centering
\caption{\normalfont Durations of the \ac{PHM} test cycles}
\label{tab:phm-test-lifes}
\begin{tabular}{@{}lr@{}}
\toprule
\textbf{Test cycle} & \textbf{Duration} \\
\midrule
Life 0 & 266150 \\
Life 1 & 1958 \\
Life 2 & 78018 \\
Life 3 & 226810 \\
Life 4 & 125349 \\
\bottomrule
\end{tabular}
\end{minipage}
\end{table}

The mean exceeds the median and lies between the 0.75 and 0.9 quantiles,
indicating strong right skew and exceptionally long upper-tail cycles. This
heterogeneity makes the task harder. Table~\ref{tab:phm-test-lifes} further
illustrates this variability: three test cycles exceed the training 0.9
quantile, whereas Life~1 is exceptionally short.

\subsection{Data Preprocessing} \label{subsec:data-preprocessing}

We apply the following preprocessing steps before evaluation. Inputs are
normalised to $[0,1]$ with \texttt{MinMaxScaler}. Of the 24 sensor signals, we
select ten following~\cite{phm_RNN_online}. \texttt{FIXTURESHUTTERPOSITION} is
a binary operational-state signal that switches from 0 to
1~\cite{phm_transformer}; we retain all input features only for samples in
which this signal equals 1, thereby considering only useful machine life. \revised{As a
result, the following sensor readings are considered: \texttt{IONGAUGEPRESSURE,
ETCHBEAMVOLTAGE, ETCHBEAMCURRENT, ETCHSUPPRESSORVOLTAGE, ETCHSUPPRESSORCURRENT,
FLOWCOOLFLOWRATE, FLOWCOOLPRESSURE, ETCHGASCHANNEL1READBACK,
ETCHPBNGASREADBACK, FIXTURESHUTTERPOSITION}.}

We use the piecewise-linear target of Section~\ref{subsec:rul-estimation},
clipping \ac{RUL} at $\text{MAX\_{RUL}}=500$. Finally, targets are normalised
to $[0,1]$ during training and denormalised for metric computation.

\subsection{\revised{Experimental Details}} \label{subsec:exp-details}

\revised{Table~\ref{tab:exp-parameters} provides additional experimental details to
ensure the reproducibility of the results.}

\begin{table}[t]
\centering
\caption{Hyperparameters and experimental settings}
\label{tab:exp-parameters}
\small
\begin{tabular}{@{}ll@{}}
\toprule
Parameter & Value \\
\midrule
\multicolumn{2}{@{}l}{\textit{Input and preprocessing}} \\
Number of input features & 9 \\
Window length / stride & 5,000 / 5,000 samples \\
Missing-value imputation & Mean imputation \\
Feature scaling & \texttt{MinMaxScaler} on $[0,1]$ \\
RUL cap / normalization & 500 / division by 500 \\
\midrule
\multicolumn{2}{@{}l}{\textit{Model architecture}} \\
$d_{\mathrm{model}} / d_{\mathrm{state}}$ / extraction layers & 128 / 64 / 4 \\
Activation / dropout & \texttt{ReLU} / 0.205 \\
\midrule
\multicolumn{2}{@{}l}{\textit{Optimization}} \\
Optimizer / initial LR / decay & \texttt{AdamW} / $10^{-3}$ / $10^{-4}$ \\
LR schedule & Cosine annealing \\
Period ($T_{\max}$) / minimum LR & 30 epochs / 0 \\
Batch size / training epochs & 8 / 30 \\
Training loss & Window-weighted Pinball loss \\
Training quantile sampling & $\tau \sim \mathcal{U}(0.1,0.9)$ \\
Evaluation quantiles & $\{0.1,0.25,0.5,0.75,0.9\}$ \\
\midrule
\multicolumn{2}{@{}l}{\textit{Validation protocol}} \\
CV folds / train--val. split / seed & 5 / 90--10\% / 42 \\
Checkpoint selection & Lowest validation pinball loss \\
\bottomrule
\end{tabular}
\end{table}

\revised{For each evaluation quantile, five runs are performed in which the model is
trained using the quantile regression framework, as explained in
Algorithm~\ref{alg:quantile-regression}, and evaluated at that specific quantile
level on the validation and test sets. This procedure yields a checkpoint for
each evaluation quantile level in each run. These checkpoints are used to
produce the results reported in Section~\ref{sec:exp-results}.}


\section{Experimental Results} \label{sec:exp-results}

This section reports the experimental evaluation of the proposed approach.
Section~\ref{subsec:all-models} compares different sequence-learning
architectures across five quantile levels; their business metrics are compared
at quantile 0.5 (i.e., the median). Section~\ref{subsec:best-model} then examines
the best-performing model across all five quantiles to demonstrate their effect
on model predictions and business metrics. To obtain robust results, we employ
five-fold cross-validation on the training cycles while holding out the test
cycles for evaluation. All results in this section are averaged across folds.

\subsection{Comparative Evaluation of Benchmark Models} \label{subsec:all-models}

This section benchmarks different \ac{DL}-based backbones for \ac{RUL}
estimation within the general pipeline described in
Section~\ref{subsec:rul-estimation}. As outlined in
Section~\ref{subsec:quantile-reg}, the \ac{QR}-based approach enables evaluation
at different quantile levels. \revised{Table~\ref{tab:all-models-quantile-tab} compares
the models across five quantiles.} Quantile 0.5 (i.e., the median) corresponds to
the well-known \ac{MAE} loss and is used below for the business-metric and
accuracy-complexity comparisons. A detailed quantile-wise evaluation of the
best model is reported in Section~\ref{subsec:best-model}.



\begin{table*}[t]
    \centering
    \caption{\normalfont Mean \ac{RMSE} $\pm$ sample standard deviation across the five cross-validation folds of the benchmark models at different evaluation quantiles. Within each fold, \ac{RMSE} is averaged over the test lives. The best result is shown in bold and the second-best in italics.}
    \label{tab:all-models-quantile-tab}
    \begin{tabular}{lccccc}
        \toprule
        \textbf{Model} & \textbf{0.1} & \textbf{0.25} & \textbf{0.5} & \textbf{0.75} & \textbf{0.9} \\
        \midrule
        S5     & 136.38 $\pm$ 22.86 & 121.71 $\pm$ 42.49 & 52.82 $\pm$ 21.73 & 48.42 $\pm$ 12.27 & \textit{47.02} $\pm$ 14.08 \\
        S4D    & \textbf{54.86} $\pm$ 32.56 & \textit{43.90} $\pm$ 34.44 & \textbf{37.17} $\pm$ 23.05 & \textbf{36.27} $\pm$ 14.51 & \textbf{36.90} $\pm$ 11.88 \\
        S4     & \textit{60.57} $\pm$ 52.74 & 58.11 $\pm$ 61.82 & 49.16 $\pm$ 44.07 & \textit{46.55} $\pm$ 31.57 & 47.86 $\pm$ 24.85 \\
        MLP    & 149.49 $\pm$ 83.40 & 162.03 $\pm$ 135.47 & 103.48 $\pm$ 40.29 & 140.31 $\pm$ 110.16 & 143.33 $\pm$ 103.28 \\
        Linear & 215.39 $\pm$ 75.14 & 297.71 $\pm$ 97.24 & 377.41 $\pm$ 200.76 & 377.66 $\pm$ 223.97 & 373.83 $\pm$ 223.41 \\
        LSTM   & 159.21 $\pm$ 103.27 & 114.68 $\pm$ 63.37 & 131.71 $\pm$ 88.65 & 133.41 $\pm$ 98.67 & 141.15 $\pm$ 91.78 \\
        RNN    & 162.25 $\pm$ 122.37 & 141.11 $\pm$ 110.43 & 107.17 $\pm$ 40.83 & 113.07 $\pm$ 45.07 & 114.65 $\pm$ 24.90 \\
        GRU    & 156.82 $\pm$ 114.70 & 129.17 $\pm$ 90.20 & 134.68 $\pm$ 112.68 & 134.27 $\pm$ 98.51 & 114.53 $\pm$ 30.40 \\
        Transformer & 67.20 $\pm$ 30.71 & \textbf{38.80} $\pm$ 22.99 & \textit{45.68} $\pm$ 35.07 & 59.55 $\pm$ 44.86 & 60.00 $\pm$ 51.00 \\
        \bottomrule
    \end{tabular}
    \vspace{0.5\baselineskip}
\end{table*}

\begin{figure}[htbp]
    \centering
    \includegraphics[width=\columnwidth]{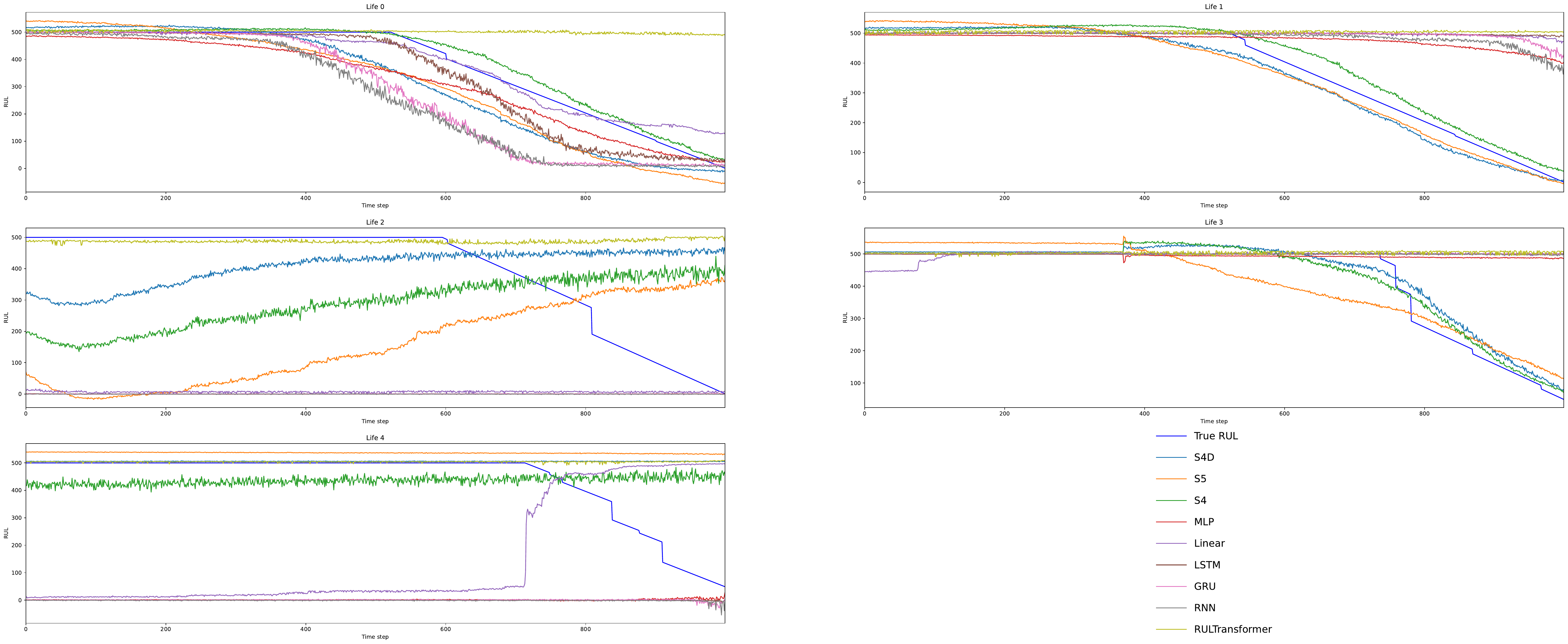}
    \caption{\ac{RUL} estimates from all benchmark models at quantile 0.5}
    \label{fig:multi-quantile-plot}
\end{figure}

\revised{Table~\ref{tab:all-models-quantile-tab} shows that the \ac{SSM}-based
architectures consistently provide the most accurate \ac{RUL} predictions
across the evaluated quantiles, with \ac{S4D} achieving the strongest overall
performance with limited variability across cross-validation folds. In
contrast, classical sequence-learning architectures and the non-sequential
\ac{MLP} and Linear baselines struggle to model the \ac{RUL} dynamics
effectively. Figure~\ref{fig:multi-quantile-plot} complements the aggregate
metrics by showing the predicted \ac{RUL} trajectories for the last 1000
samples of each test life. The \ac{SSM} models are the only approaches that
consistently respond to the degradation phases: \ac{S4D} closely follows the
decline in Life~1 and captures the late-life decreases in Lives~0 and~3, while
\ac{S4} and \ac{S5} also reproduce parts of these trends. Their errors are
nevertheless apparent on Life~2 and Life~4, where the predicted trajectories
depart markedly from the target after the decline begins. In contrast, the
\ac{MLP}, Linear, \ac{LSTM}, \ac{GRU}, \ac{RNN}, and Transformer baselines
remain nearly constant for substantial portions of Lives~1, 3, and~4 or collapse
prematurely in Life~2, and therefore fail to represent the observed \ac{RUL}
evolution. This trajectory-level comparison is consistent with the superior
overall accuracy of \ac{S4D} in Table~\ref{tab:all-models-quantile-tab}.}


\begin{figure}[htbp]
    \centering
    \includegraphics[width=\columnwidth]{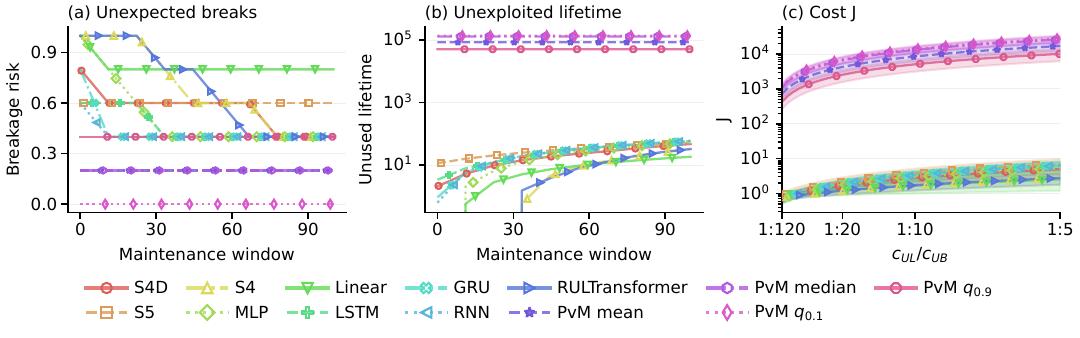}
    \caption{Business metrics across the benchmark models}
    \label{fig:business-metrics-models}
\end{figure}

%
%


\revised{Figure~\ref{fig:business-metrics-models} evaluates the maintenance decisions
obtained from the \ac{DL} models as the maintenance-window size increases. In
this evaluation, \ac{PdM} models are compared with \ac{PvM} baselines, denoted
by \texttt{mean}, \texttt{median} and the 0.1 and 0.9 extreme quantile levels.
Their maintenance points are computed from the mean, median, and 0.1 and 0.9
quantiles of training-cycle durations, respectively. The left panel
(Fig.~\ref{fig:business-metrics-models}a) reports $\rho_{UB}$, namely the
fraction of test lives that experience an unexpected break before the planned
intervention; the centre panel (Fig.~\ref{fig:business-metrics-models}b)
reports $\rho_{UL}$, i.e., the \ac{UL} associated with an early intervention.
Finally, the right panel (Fig.~\ref{fig:business-metrics-models}c) reports a
sensitivity analysis of $J$, defined by~\eqref{eq:metric-J}: its horizontal
axis is the cost ratio $c_{UL}/c_{UB}$, while its vertical axis shows the
corresponding $J$ values.} This ratio expresses the cost of one unit of
\ac{UL} relative to the cost of one \ac{UB}. We use this ratio-based
representation because the absolute values of $c_{UB}$ and $c_{UL}$ must be
determined for the specific application through a careful cost study, which
should identify the pair that minimises $J$. Lower values of $J$ indicate a
more favourable maintenance trade-off. As the maintenance window increases, the
\ac{PdM} curves reduce the risk of unexpected breaks because maintenance is
requested earlier; this reduction is accompanied by the expected increase in
unexploited lifetime. In contrast, the preventive baselines maintain nearly
constant, exceptionally large \ac{UL} values, reflecting their systematically
early interventions induced by the right-skewed training-life distribution.
This behaviour is also evident in the $J$ panel: despite their low or null
breakage risk, the large amount of discarded life makes the baseline costs
orders of magnitude higher than those of the \ac{PdM} models.


\begin{figure*}[t]
    \centering
    \subfloat[Model parameters \label{fig:pareto-rmse-parameters}]{%
        \includegraphics[width=0.40\textwidth]{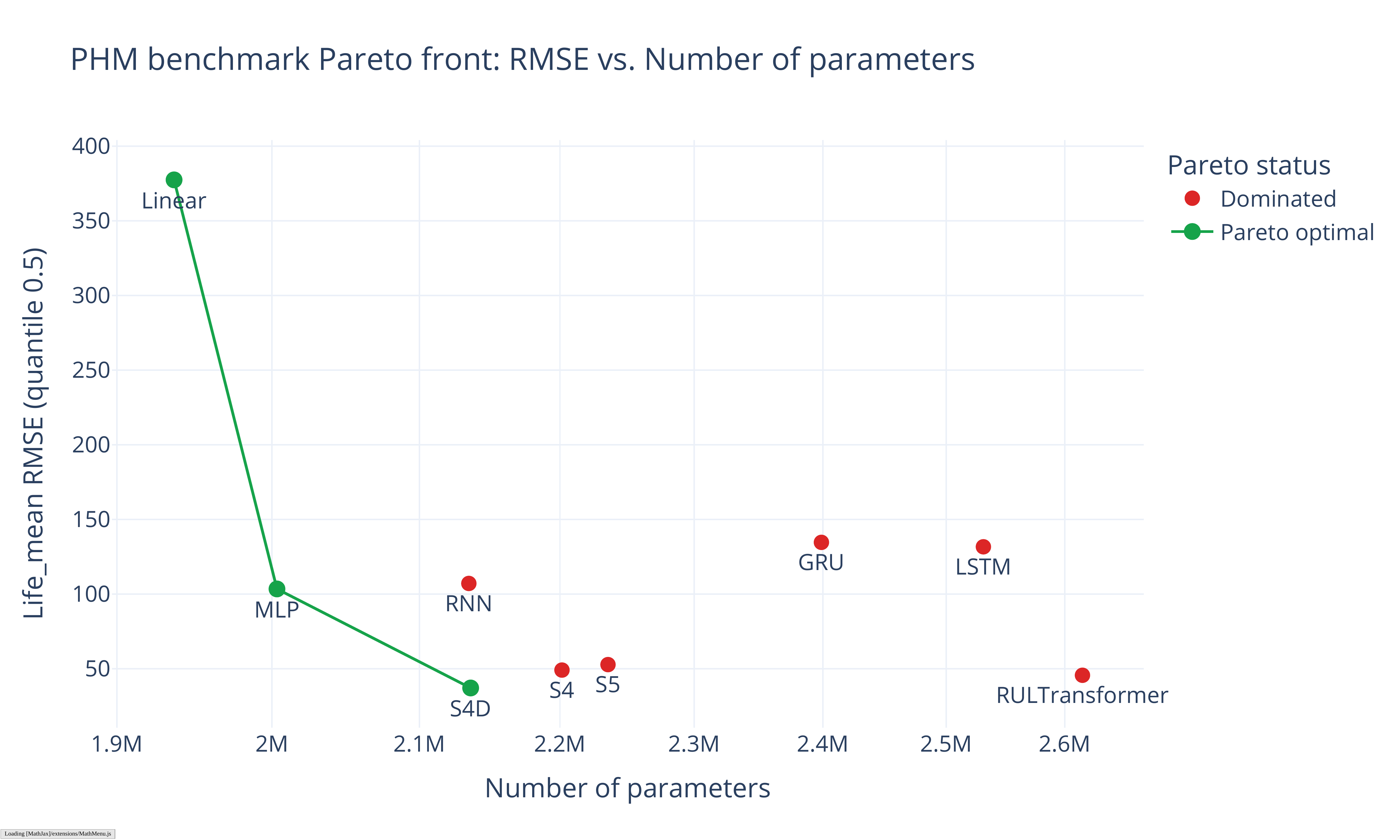}
    }\hspace{0.025\textwidth}%
    \subfloat[GFLOPs \label{fig:pareto-rmse-gflops}]{%
        \includegraphics[width=0.40\textwidth]{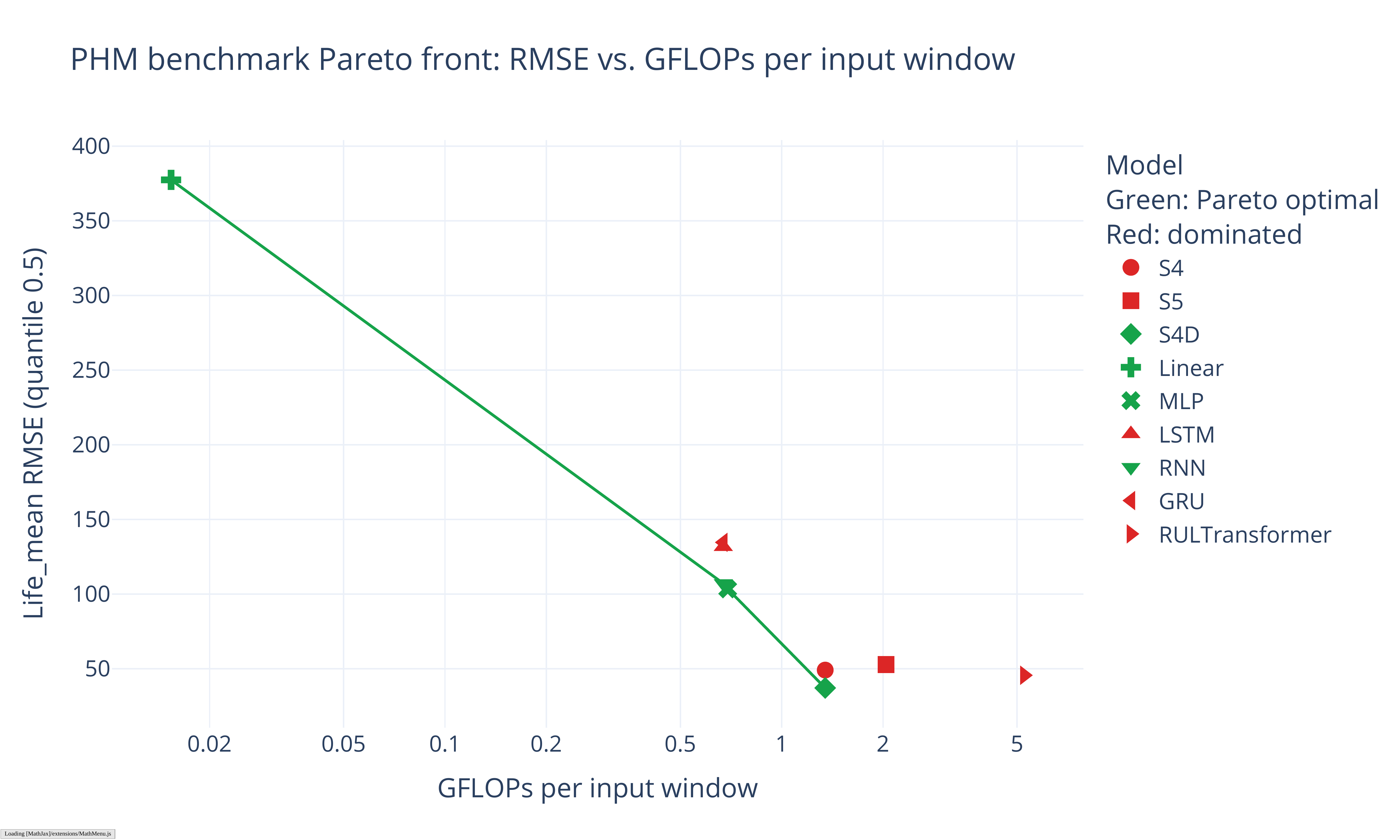}
    }

    \caption{Predictive accuracy--complexity trade-off across benchmark models}
    \label{fig:pareto-fronts}
\end{figure*}

\revised{Figure~\ref{fig:pareto-fronts} compares predictive accuracy with model
complexity (i.e., the number of model parameters in the left panel) and
computational cost (i.e., GFLOPs in the right panel) at quantile 0.5. \ac{S4D}
achieves the lowest
\ac{RMSE} with roughly two million parameters.} More importantly, it lies on
the Pareto front in both panels: no competing model simultaneously improves its
prediction error and its parameter count or computational cost. This confirms
that \ac{S4D} combines high predictive accuracy with both a small memory
footprint and high throughput. The closely related \ac{S4} and \ac{S5} models
also yield low errors, whereas the \ac{MLP}, \ac{RNN}, \ac{GRU}, and \ac{LSTM}
have substantially higher errors despite comparable or larger resource
requirements. \revised{The Linear model is also present on both Pareto fronts because
its parameter count and computational cost cannot be matched by
the other models.} Its prediction error, however, is too high for practical
application; its Pareto optimality therefore reflects its exceptionally low
resource usage rather than a useful accuracy-efficiency compromise. At the
opposite extreme, the Transformer achieves accuracy comparable to that of the
\ac{SSM} models but with prohibitively high complexity. Overall, the
\ac{SSM}-based models, especially \ac{S4D}, provide the most favourable and
practically relevant accuracy-complexity trade-off in this benchmark.

\subsection{Quantile-wise Analysis of the Best-Performing Model} \label{subsec:best-model}

This section considers the performance of \ac{S4D}, the best-performing model
according to Table~\ref{tab:all-models-quantile-tab}, at all evaluation
quantiles to demonstrate how modelling uncertainty through \ac{QR} shapes
model behaviour.


\begin{table}[t]
    \centering
    \caption{\normalfont \ac{RMSE} of \ac{S4D} at different evaluation quantiles}
    \label{tab:S4D-metrics-table}
    \begin{tabular}{lccccc}
        \toprule
        \textbf{Test cycle} & \textbf{0.1} & \textbf{0.25} & \textbf{0.5} & \textbf{0.75} & \textbf{0.9} \\
        \midrule
        Life~0 & 19.77 & 15.56 & 12.49 & 11.66 & 12.35 \\
        Life~1 & 184.32 & 142.72 & 117.00 & 114.81 & 117.87 \\
        Life~2 & 32.24 & 31.74 & 30.30 & 29.50 & 29.28 \\
        Life~3 & 19.63 & 11.15 & 6.65 & 6.01 & 5.36 \\
        Life~4 & 18.36 & 18.31 & 19.42 & 19.40 & 19.62 \\
        \midrule
        Mean   & 54.86 & 43.90 & 37.17 & 36.27 & 36.90 \\
        Median & 28.43 & 24.65 & 23.09 & 23.91 & 24.14 \\
        Std     & 60.40 & 47.90 & 38.34 & 37.12 & 38.18 \\
        \bottomrule
    \end{tabular}
    \vspace{0.5\baselineskip}
\end{table}

Table~\ref{tab:S4D-metrics-table} shows how the effect of the evaluation
quantile varies across test lives. The metrics generally improve as the quantile
increases for Lives 0, 2, and 3, while the results for Life 4 remain nearly
constant across quantiles. In contrast, the model performs less accurately on
Life 1 at every quantile. As discussed in Section~\ref{sec:exp-setup}, its exceptionally
short duration differs substantially from the predominantly long and highly
variable training life cycles, making it difficult for the model to generalise
to this case.


\begin{figure}[htbp]
    \centering
    \includegraphics[width=\columnwidth]{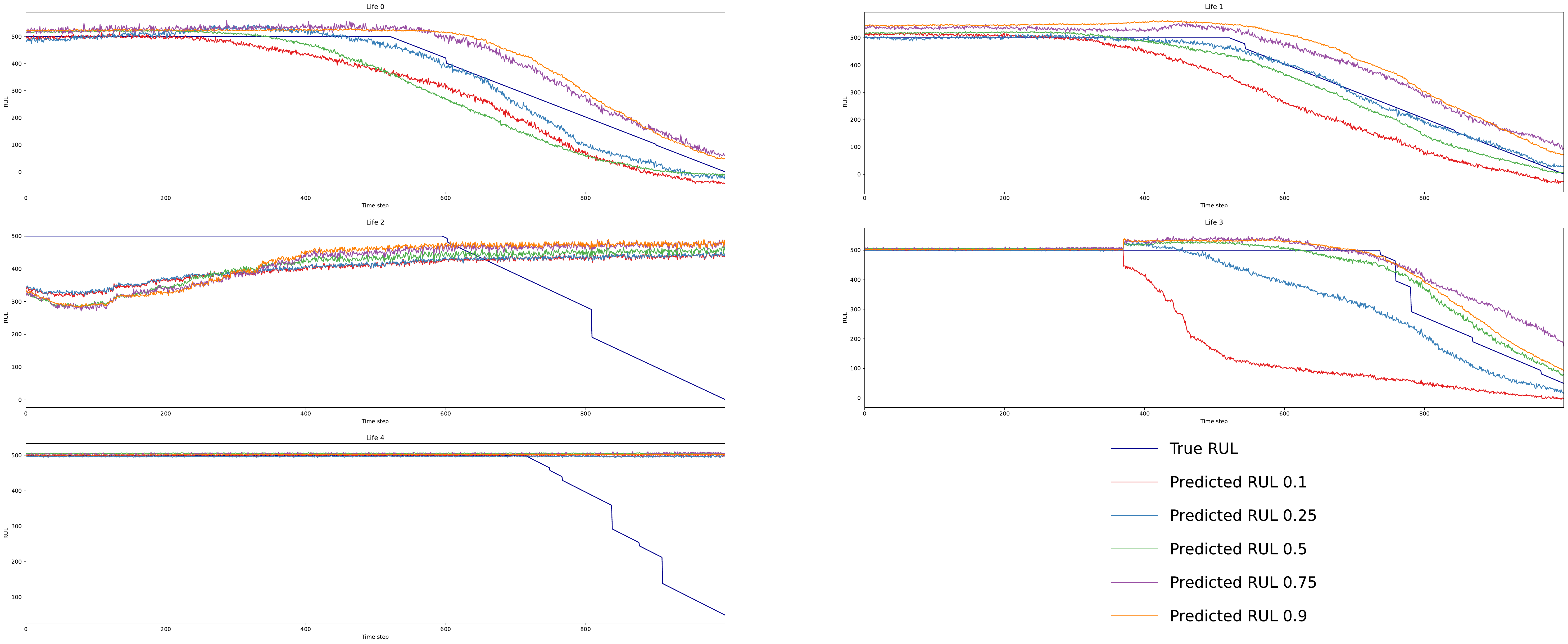}
    \caption{\ac{S4D} \ac{RUL} estimates across evaluation quantiles}
    \label{fig:S4D-quantile-plot}
\end{figure}

Figure~\ref{fig:S4D-quantile-plot} shows that the quantile estimates generally
capture the declining trend of the \ac{RUL} for Lives 0 and 1, with lower
quantiles providing more conservative predictions and upper quantiles yielding
progressively larger \ac{RUL} estimates. The separation between quantile
trajectories becomes more pronounced as degradation progresses, reflecting
increased predictive uncertainty near failure. Performance is less consistent
for Lives 2, 3, and 4: the model does not reproduce the abrupt terminal decrease
in Lives 2 and 4 and exhibits substantial quantile-dependent bias for Life 3.
These results are probably due to the skewed life distribution described in
Section~\ref{par:data-description}, particularly the exceptionally short Life
1.


\begin{figure}[htbp]
    \centering
    \includegraphics[width=\columnwidth]{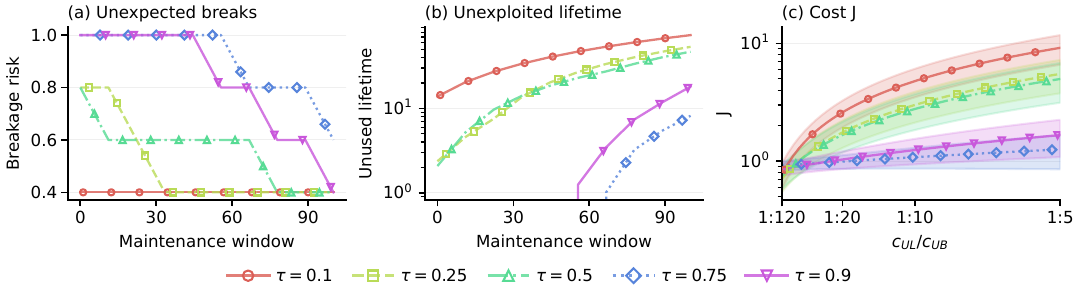}
    \caption{Business metrics across evaluation quantiles for the \ac{S4D} model}
    \label{fig:quantile-business-metrics-S4D}
\end{figure}

%
%

\revised{Figure~\ref{fig:quantile-business-metrics-S4D} isolates the effect of the
evaluation quantile on the maintenance trade-off for \ac{S4D}. For the higher
quantiles, 0.75 and 0.9, the risk of \ac{UB} initially remains high, but
decreases sharply as the maintenance window is enlarged. As expected, the
\ac{UL} increases with the maintenance window for every quantile, since
interventions are scheduled earlier. However, the 0.75 and 0.9 quantiles
consistently yield lower \ac{UL} than the lower quantiles. This reduced \ac{UL}
is also reflected in the $J$ curves, where these two quantiles achieve the
lowest costs over the examined cost ratios, despite their higher initial
\ac{UB} risk. Thus, for the cost scenarios considered here, the savings
associated with avoiding excessively early maintenance outweigh the
corresponding increase in breakage risk.}



\section{Conclusion} \label{sec:conclusion}

We presented a \ac{PdM} framework for ion-milling \ac{RUL} estimation that
combines \ac{DL} sequence models and \ac{SQR} to estimate conditional
quantiles. Quantile selection adjusts maintenance decisions to operational
risk, while \ac{UB}, \ac{UL}, and cost-weighted $J$ assess their business
consequences and identify a failure-utilisation trade-off. This solution is
particularly useful in the \revised{context} of semiconductor manufacturing, where making
risk-informed decisions is of paramount importance to prevent a single failure
from disrupting the entire production pipeline. On the $F_1$ \ac{PHM18} fault mode,
\ac{SSM} architectures were the most efficient in terms of memory and computation and produced
the most accurate \ac{RUL} estimates, with \ac{S4D} best overall across
quantiles. Quantile and maintenance-window choices controlled the \ac{PdM}
trade-off: larger windows reduced \ac{UB} for \ac{PdM} models while gradually
increasing \ac{UL}. By contrast, \ac{PvM} baselines incurred very large \ac{UL}
because their decisions used summary statistics of a strongly right-skewed
training-life distribution. Their $J$ values were therefore much higher despite
low breakage risk, showing that quantile-based \ac{RUL} estimation can support
more efficient, risk-aware policies.

\revised{Future research may extend the approach beyond the $F_1$ fault mode by
transferring knowledge learned from the available data to other failure types
in \ac{PHM18} (i.e., $F_2$ and $F_3$), for example through domain-adaptation
techniques~\cite{domain_adaptation}.} Continual learning strategies could also
further enable the model to adapt to new fault conditions and evolving
operating regimes without repeatedly retraining it from scratch. Additional
directions include evaluating the framework on larger and more diverse
industrial datasets, calibrating the quantile estimates and maintenance-cost
parameters with plant-specific operational data, and studying adaptive policies
that jointly select the quantile level and maintenance window online.

\bibliographystyle{IEEEtran}
\bibliography{
  refs/related_work,
  refs/ssm,
  refs/quantile,
  refs/monotonic
}




\end{document}